# YOLOv5s-GTB: light-weighted and improved YOLOv5s for bridge crack detection

Xiao Ruiqiang[a]

[a]Southeast University, Nanjing, 211189, China

Abstract

In response to the situation that the conventional bridge crack manual detection method has a large amount of human and material resources wasted, this study is aimed to propose a light-weighted, high-precision, deep learning-based bridge apparent crack recognition model that can be deployed in mobile devices' scenarios. In order to enhance the performance of YOLOv5, firstly, the data augmentation methods are supplemented, and then the YOLOv5 series algorithm is trained to select a suitable basic framework. The YOLOv5s is identified as the basic framework for the light-weighted crack detection model through experiments for comparison and validation.By replacing the traditional DarkNet backbone network of YOLOv5s with GhostNet backbone network, introducing Transformer multi-headed self-attention mechanism and bi-directional feature pyramid network (BiFPN) to replace the commonly used feature pyramid network, the improved model not only has 42% fewer parameters and faster inference response, but also significantly outperforms the original model in terms of accuracy and mAP (8.5% and 1.1% improvement, respectively). Luckily each improved part has a positive impact on the result.

This paper provides a feasible idea to establish a digital operation management system in the field of highway and bridge in the future and to implement the whole life cycle structure health monitoring of civil infrastructure in China.

*Keywords:* Bridge cracks, object detection, YOLOv5, GhostNet, Attention mechanism

# 1 Introduction

Transport has gradually become an important factor affecting the living standards of the masses as China's economic strength continues to grow. With the increase in people's desire for travel, the total mileage of highway bridges has been increasing. According to statistics, the total mileage of highway in the country has reached 457.73 million kilometers by the year 2021, including 780 million highway bridges, measuring about 460 million meters, as well as 3894 very large bridges measuring about 694.20 million meters[1]. A majority of them are small bridges made of concrete, most of which have been in service for 10 to 20 years. During construction and operation, concrete structures may crack due to internal and external influences such as temperature, shrinkage, or foundation deformation[2].

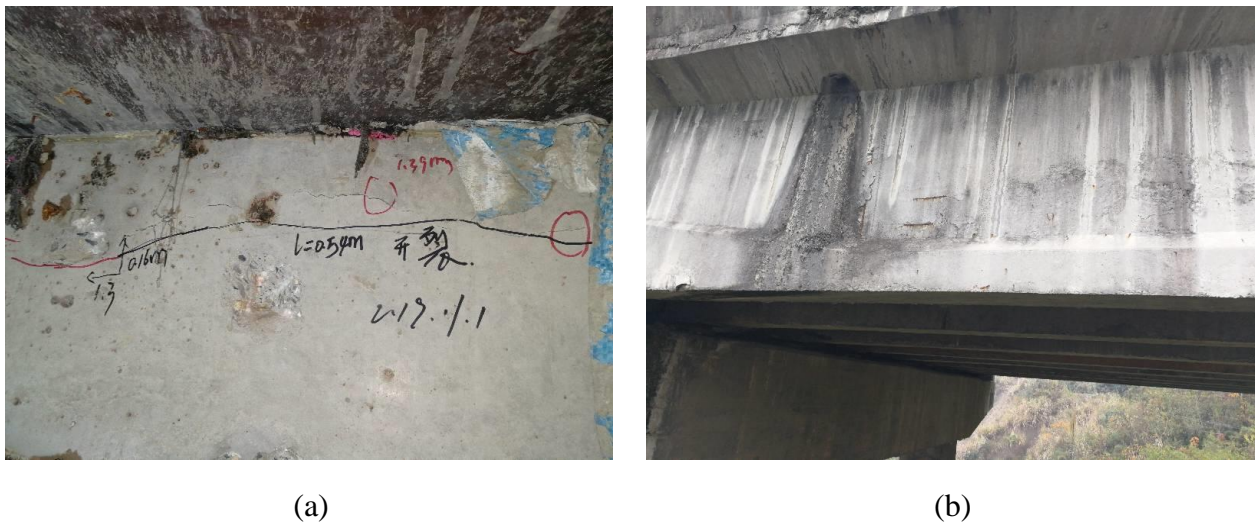

(a)　　　　　　　　　　　　　　　(b)

Figure 1-1

Cracks in bridges usually appear at the microscopic level on the surface of infrastructure components[3]. As a result of cracks, components lose their load-bearing capacity, and surface discontinuities develop[4–6]. The further damage can be reduced if cracks of this type are detected early on. Undetected cracks, however, can spread to the entire surface and result in a collapse, resulting in fatalities, injuries, and economic loss. Therefore, concrete infrastructure should be monitored throughout its life cycle and should be complemented by a traceable digital operation and maintenance system to continuously monitor internal and external changes on the structure.

Traditional manual crack detection methods have many problems: they require a lot of human, material and financial resources; they are inefficient due to the lack of control over the professional quality of the inspectors; they are susceptible to external environmental factors; and they cannot guarantee the safety of the inspectors. Therefore, some researchers are considering the possibility of using computer arithmetic to "replace" the human eye. So the research method of crack detection technology based on deep learning has been gradually developed. The use of deep learning and image processing technology together to complete the task of crack recognition can improve the efficiency of detection, reduce the cost of detection, and ensure the safety of inspectors.

In recent years, with the rapid development and grounded application of deep learning technology, the feasibility of infrastructure crack recognition model based on computer vision has been greatly improved. Deep convolutional neural network models trained based on a large amount of raw data, together with image data enhancement technology, can to a certain extent complete the task of automatic detection and classification of infrastructure cracks or other diseases. If this method is implemented in reality, it can not only solve various problems caused by the human eye, but also greatly improve the speed and accuracy of detection, which has great research and application value.

Convolutional Neural Network (CNN) models are commonly used in the literature of crack detection. The model consists of three layers of neurons: a convolutional layer, a pooling layer, and a fully connected layer. The convolutional layer extracts features from the image, enabling it to learn to distinguish between cracked and non-cracked images. The pooling layer is used to downsample the image and reduce its size by resizing it. the final stage of the CNN model uses the fully connected layer because it takes the output of the previous layer as input and maps it to the output labels. Ni et al. automated these tasks using feature graph fusion and pixel classification by applying a CNN architecture called GoogleNet to crack classification [10]. A feature pyramid network (FPN) is used in this method to process the output, in which the network contains feature fusion layers and successive convolutional layers to characterize the cracks at multiple scales. The results show that the network is able to describe the cracks accurately with an accuracy of 80.13%. However, at the same time, the large resolution of the crack images leads to long processing time of the neural network, which takes about 16 seconds to detect in a 6000×4000 pixel image.

The Region-Convolutional Neural Network (R-CNN) structure has been applied in the field of crack detection [12], in which selective search was first used to detect candidate regions of cracks. Then, crack classification and bounding box regression were performed using CNNs pre-trained on ImageNet and Cifar-10 datasets.Li et al [13] designed a multi-scale defective region suggestion network (RPN) that proposed candidate bounding boxes in different layers, thus improving the detection accuracy. They also performed crack geolocation using an additional depth architecture and a geotagged image database. It is noteworthy that both the crack detection network and the geolocation module are incorporated into one network. An improved version of a fast R-CNN in the field of crack detection, called CrackDN [14], has been proposed among foreign scholars. In this framework, a pre-trained CNN is used to extract deep features and another sensitivity detection network to speed up the training, and the detection targets of this network include images with unbalanced illumination of cracks and shadows.

# 2 Related Work

## 2.1 Object detection

With the continuous improvement of computer computing power and model complexity, computer vision has the tendency to gradually replace the human eye. Several basic recognition problems in computer vision are currently image classification [18], target detection, semantic segmentation [12,19], and instance segmentation [20] (see Figure 2-1). Among them, the task of image classification (Figure 2-1(a)) is that given an image, the computer needs to identify the semantic class of the target object. In contrast to image classification, target detection requires not only identifying the target object class, but also predicting the location of the Bounding Box for each target (Figure 2-1(b)). In general, target detection consists of many subtasks, including face recognition [21], pedestrian recognition [22], and skeleton recognition [23].

Semantic segmentation (Figure 2-1 (c)) aims at providing and predicting pixel-level classifiers and assigning a specific class label to each pixel, thus providing a richer understanding of the image. However, in contrast to target detection, semantic segmentation does not distinguish between multiple target objects of the same class. For this reason, a combination of target detection and semantic segmentation has been proposed and named instance segmentation - which not only recognizes different targets but also assigns pixel-level masks for each class of objects.

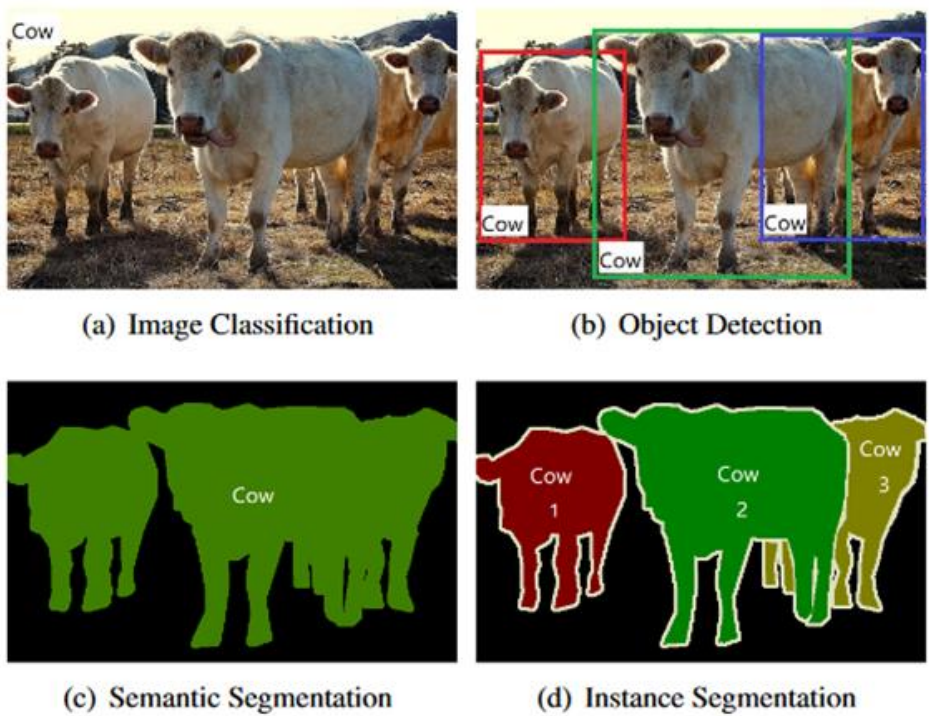

Figure 2-1 Computer vision tasks[24]

## 2.2 Evaluation of CNN model

### 2.2.1 Precision and Recall

Precision and recall are widely used in machine learning and deep learning to measure the accuracy of classifiers in binary or multiclassification problems. In binary classification problems, the Confusion Matrix is often used to represent the relationship between the positive and negative predicted values and the positive and negative true values.

The calculation of P and R is shown below：

$$P = \frac{TP}{TP+FP}$$
$$R = \frac{TP}{TP+FN}$$

### 2.2.2 Mean Average Precesion (mAP)

In multi-target detection, Average Precesion is the area under the average P-R curve for a uniform selection of 11 different recall rates, and is the integral of precision over recall.

(Mean Average Precesion) is usually used as a measure of the performance of the target detection model, and is the average of the average precision under different detection categories, i.e., the sum of the average precision of all categories divided by the number of all categories, i.e., the average of the average precision of all classes in the data set, assuming that the number of classes in the data set is generally equally distributed. The following equation is the formula for finding the

$$mAP = \frac{1}{M}\sum_{i=1}^{M} AP_i$$

### 2.2.3 Floating point operations (FLOPs)

When measuring the complexity of a deep learning algorithm or model, comparisons are often made using the number of floating point operations (FLOPs) - the amount of computation a model undergoes during operation. The experimental procedure in the second half of this paper uses units specifically $GFLOPs(giga\ FLOPs)$, where $1GFLOPs = 10^9\ FLOPs$ 。

# 3 YOLOv5s-GTB

## 3.1 Overview of YOLOv5s-GTB

Considering the daily bridge maintenance and overhaul aspects involved in this paper, not only the model training duration needs to be able to meet the daily inspection frequency requirements, but also it is best to be deployed on mobile devices of construction monitors, such as cell phone tablet drones, etc. Therefore, for the task of bridge apparent crack identification, how to design a deep learning model that can meet both the accuracy requirements of daily bridge inspection and the low recognition rate and light weight becomes the focus of this chapter. This chapter focuses on the process of building a lightweight crack detection model based on YOLOv5 improvements. The main contents of this chapter include the following four major parts: data set preparation, data enhancement improvement, changes to the YOLOv5 network structure for the lightweight goal, and analysis of the results of the ablation test.

## 3.2 Data augmentation

### 3.2.1 Basic manipulation

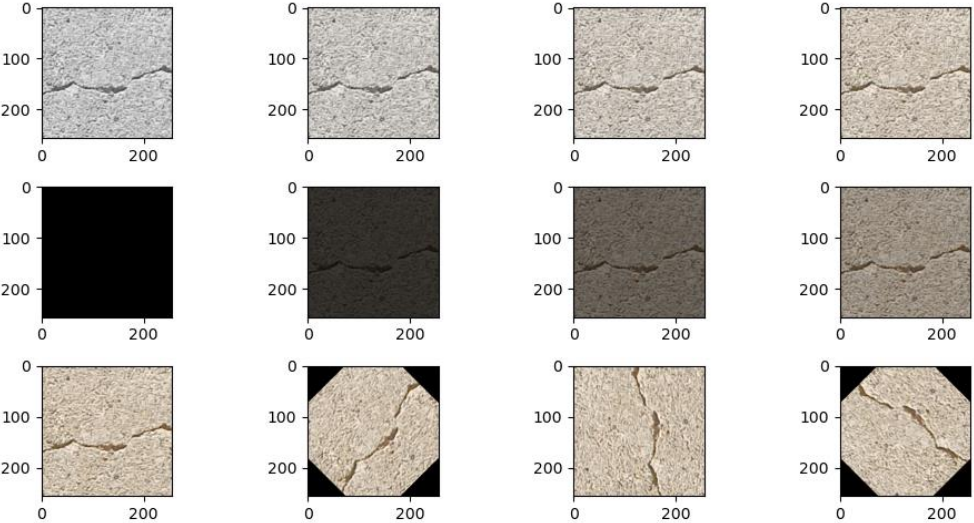

（a）

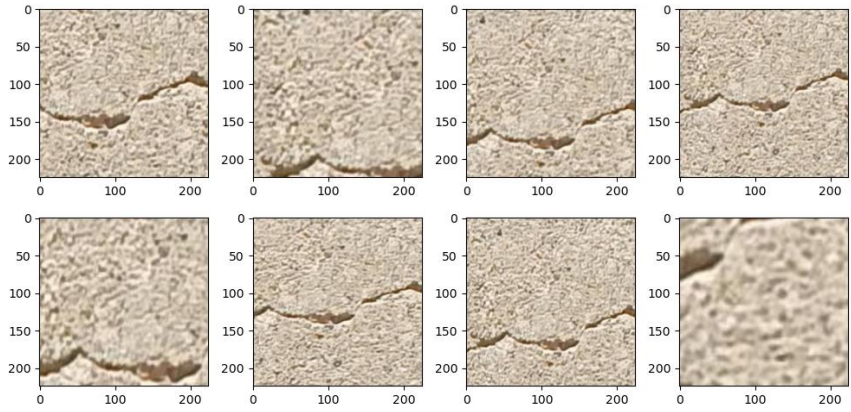

（b）

Figure 3-2

## 3.2.2 Random erasing

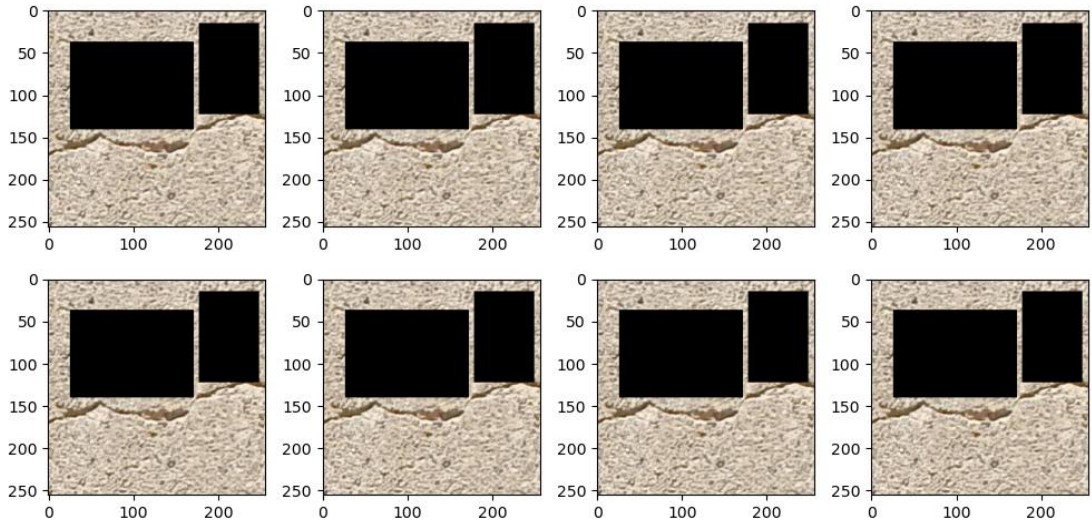

Figure 3-3

## 3.3 YOLOv5s-GTB

### 3.3.1 GhostNet Backbone

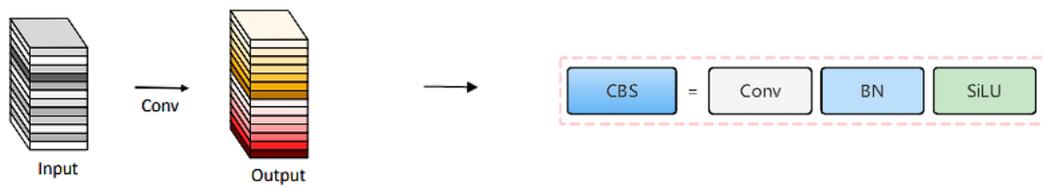

(a) Conv

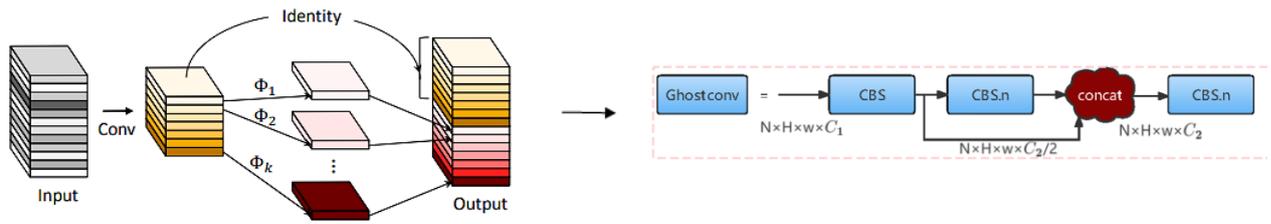

(b) GhostConv

Figure 3-10

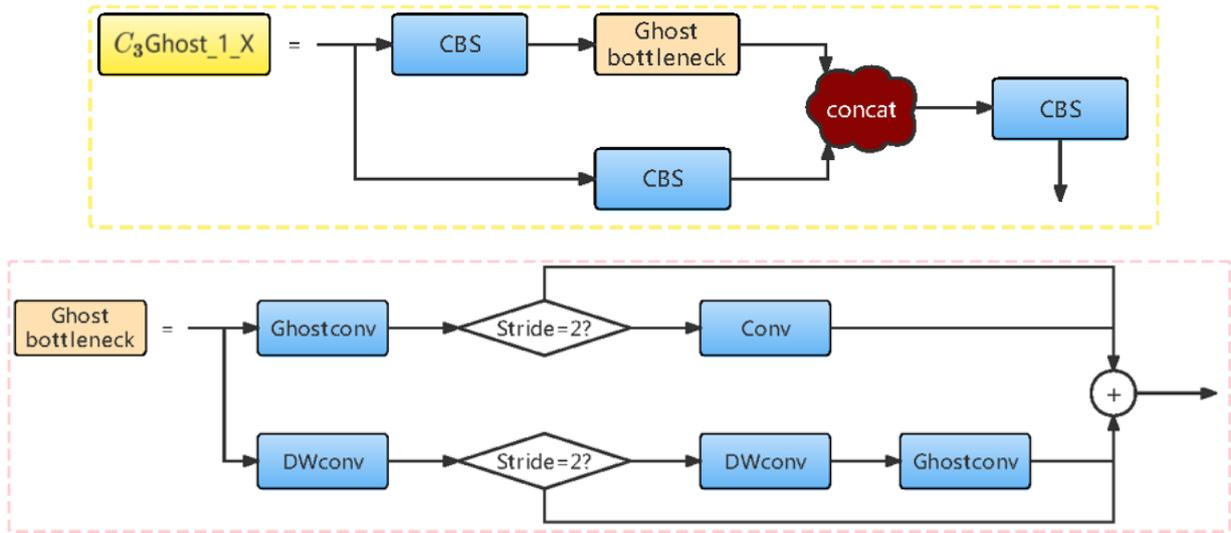

Figure 3-11

## 3.3.2 Transformer Layer

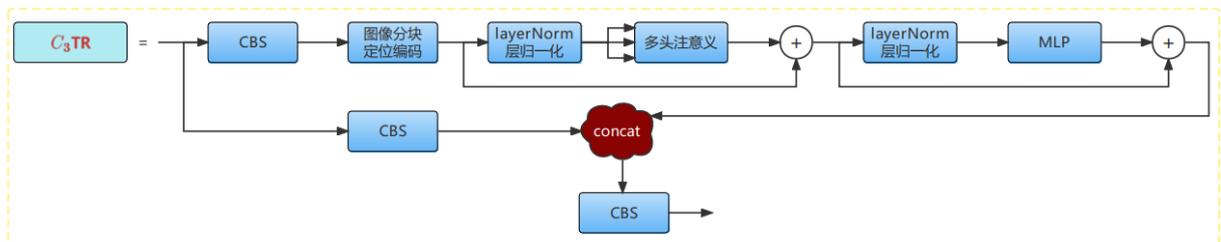

Figure 3-13

With the addition of the attention module, the features are channel-adjusted with a small increase in the number of operations and model complexity, under which useful information is given greater retention weight and useless information is suppressed to enhance the feature extraction capability of YOLOv5s while maximizing the reduction of the detrimental effects due to the decreasing number of GhostNet parameters.

### 3.3.3 BiFPN

Small-scale targets are relatively ineffective in the current mainstream target detection algorithms. With the gradual application of multiscale feature fusion to the feature map generation enhancement stage, the accuracy of small-scale target recognition can be significantly improved. the FPN feature fusion approach used by YOLOv5s and the bi-directional feature pyramid network (BiFPN) [47] feature fusion approach introduced in this section are both more popular methods to solve small-scale target detection.

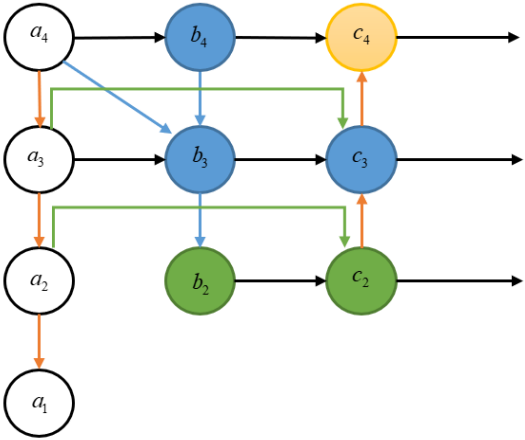

Figure 3-11

### 3.3.4 Overall network structure

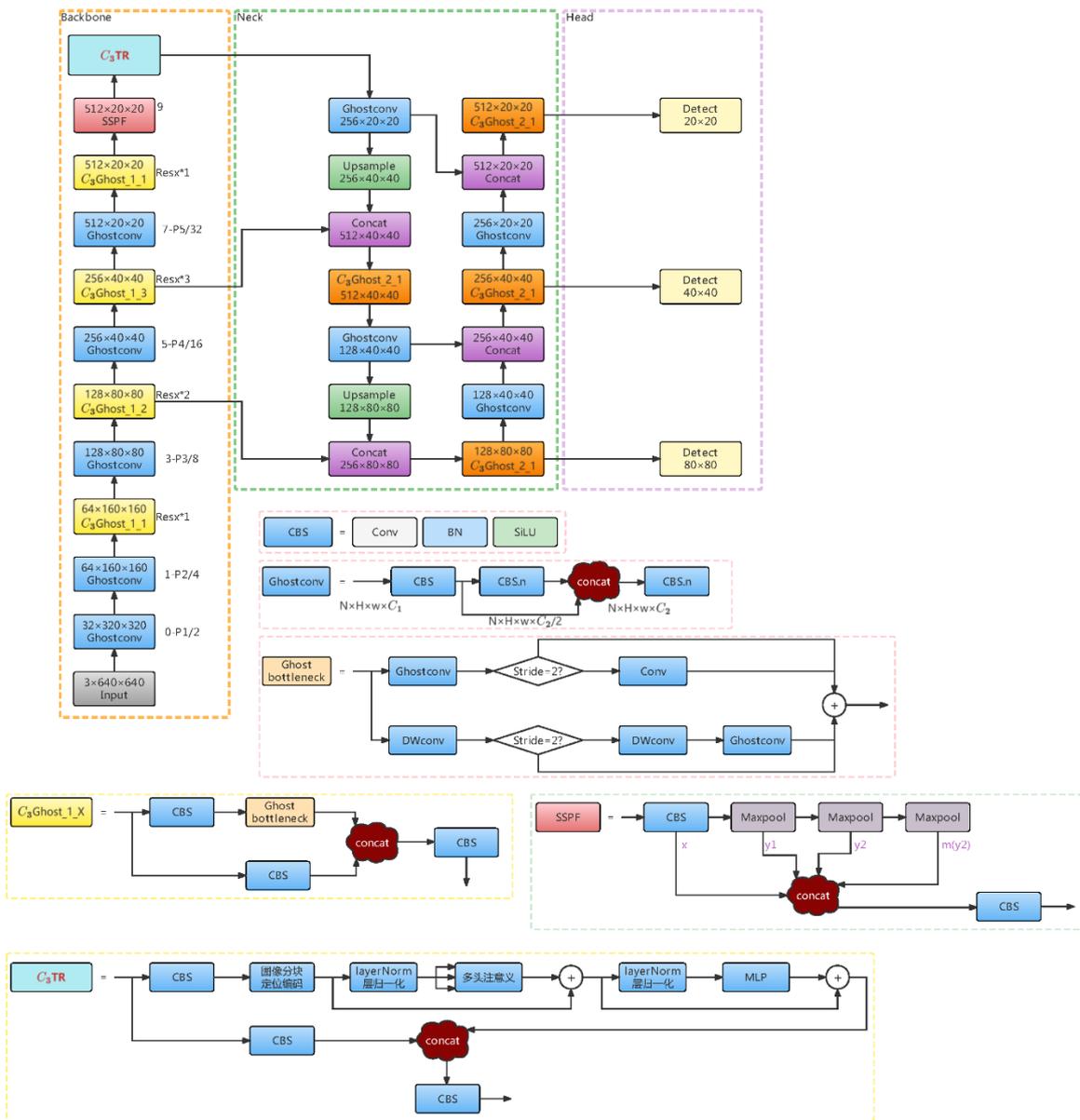

# 4 Numerical experiments

This section uses PyTorch framework to build the base network with 1 Nvidia Geforce RTX 3080 Laptop GPU, 16384MB of graphics memory, 1 Intel i7-125600H CPU, 32GB of overall memory, with Pytorch Version-1.10.0 and NVIDIA cuDNN (neural network GPU acceleration library) Version-11.3.

## 4.1 crack dataset detection

|  | P | R | mAP | Number of parameters | GFLOPs |
|---|---|---|---|---|---|
| YOLOv5s | 0.857 | **0.875** | 0.885 | 7012822 | 15.8 |
| YOLOv5s-GTB | **0.93** | 0.833 | **0.895** | **4923214** | **9.1** |

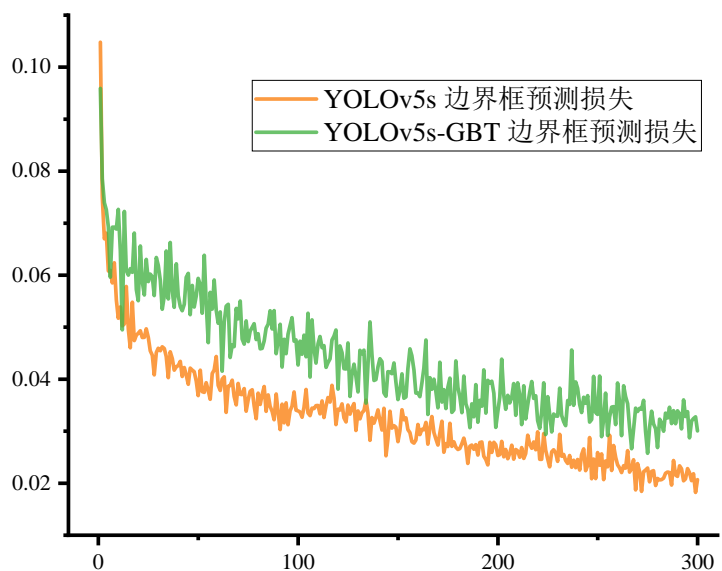

(a)

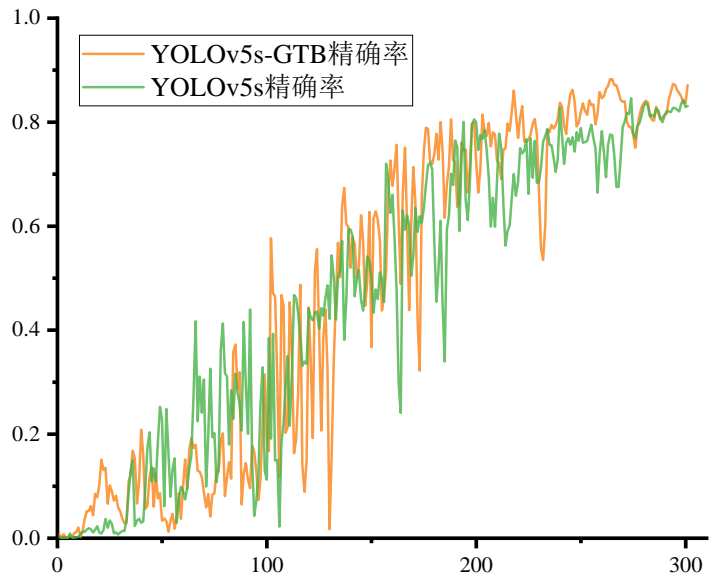

(b)

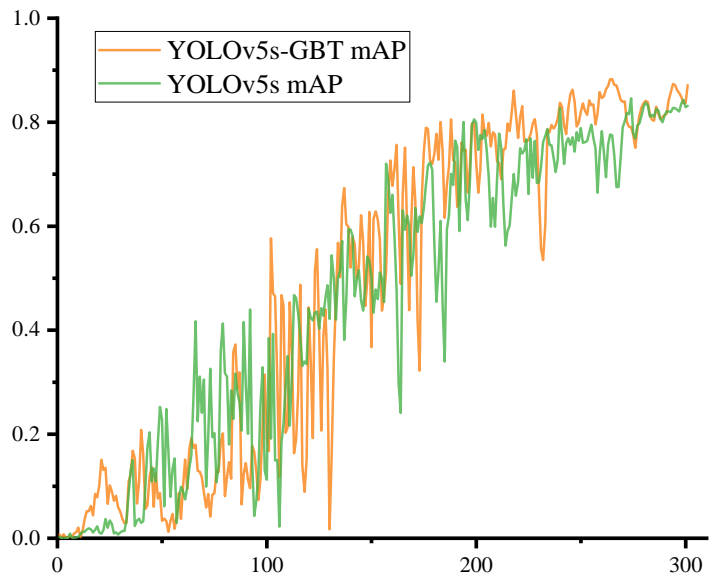

(c)

Figure 4-1

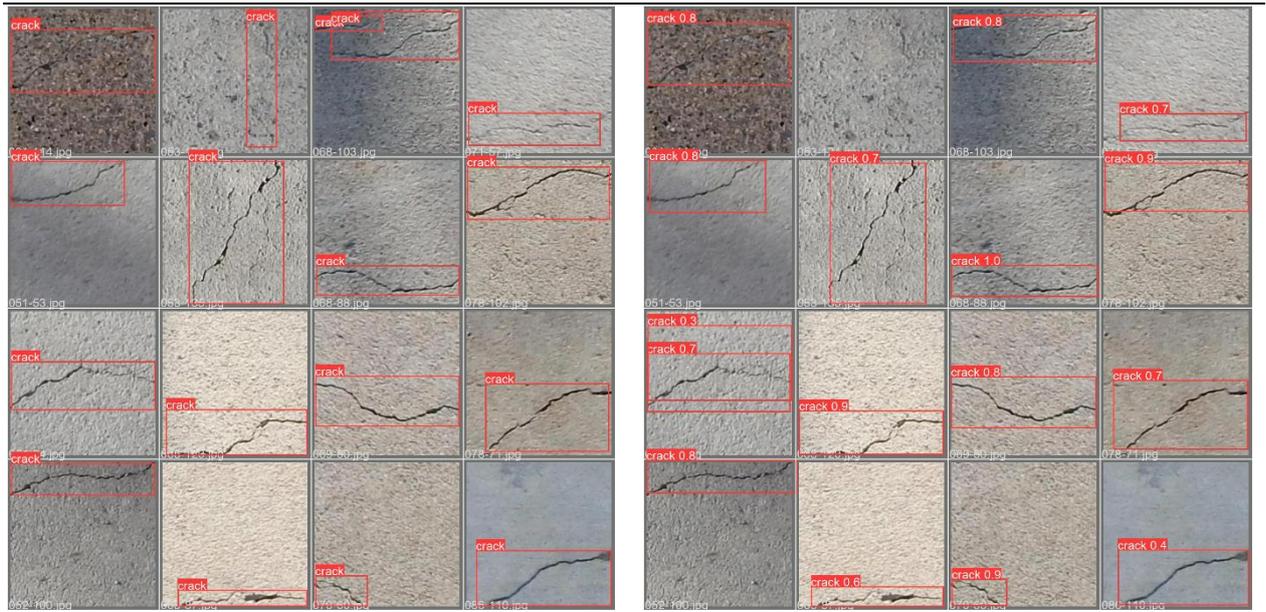

## 4.2 Ablation Studies

| Basic model | Updated Part | | | | P | R | mAP | Number of parameters | GFLOPs |
| --- | --- | --- | --- | --- | --- | --- | --- | --- | --- |
| YOLOv5s | G | T | CA | B | | | | | |
| ✓ | | | | | 0.857 | **0.875** | 0.885 | 7012822 | 15.8 |
| ✓ | ✓ | | | | 0.784 | 0.833 | 0.826 | 3675726 | **8.1** |
| ✓ | | ✓ | | | 0.913 | 0.688 | 0.83 | 7013590 | 15.6 |
| ✓ | | | ✓ | | 0.927 | 0.791 | 0.864 | 7037430 | 15.9 |
| ✓ | | | | ✓ | 0.894 | 0.706 | 0.833 | 7078358 | 16.1 |
| ✓ | ✓ | ✓ | | | 0.91 | 0.854 | 0.881 | 4857678 | 8.9 |
| ✓ | ✓ | | ✓ | | 0.974 | 0.812 | 0.856 | 3700334 | **8.1** |
| ✓ | | ✓ | | ✓ | 0.752 | 0.708 | 0.756 | 7013590 | 15.8 |
| ✓ | ✓ | | ✓ | ✓ | **0.945** | 0.729 | 0.866 | **3765870** | 8.4 |
| ✓ | ✓ | ✓ | | ✓ | 0.93 | 0.833 | **0.895** | 4923214 | 9.1 |

In order to verify the effectiveness of each part of the model, ablation experiments are conducted on YOLOv5s-GTB.

GhostNet backbone network: In this paper, we decided to introduce GhostConv as an alternative to regular convolution to extract features, in order to achieve the requirements of reducing the model network parameters, lowering the complexity of operations and lightweighting. ghostNet backbone network can effectively reduce the number of model parameters and floating point operations (see Table 3-7).

Transformer self-attentive layer: Based on the GhostConv reduced model, the Transformer self-attentive layer is used to aggregate and optimize the obtained features, so as to obtain higher detection accuracy and maximize the retention of effective information of the features. Experiments show that the Transformer self-attentive layer helps the model to improve 0.4-5.5% mAP (see Tables 3-7).

Neck part of BiFPN: Through effective bi-directional cross-scale connectivity and weighted feature fusion, the degree of contribution of different resolutions is effectively distinguished, and the extraction efficiency of more important features will be significantly improved, and the fusion process will have a higher probability of being retained, effectively removing noise and redundant information while improving accuracy. This module helps the model to improve 0.1% mAP (see Table 3-7).

# 5 Conclusion

In response to the traditional conventional bridge crack detection methods, which are characterized by a large amount of wasted human and material resources and a tedious process requiring expert confirmation, this study attempts to propose a lightweight, high-precision, bridge apparent crack recognition model that can be deployed in mobile terminal scenarios. Deep learning based target detection methods from CNN to SSD, from RCNN to the recent ViT Transformer have been developed maturely, but considering the limitations such as time cost, computational complexity, and the need for a large number of samples and labels, building a lightweight deep learning crack detection model becomes the focus of this research paper. This will have a positive effect on the implementation of whole life cycle monitoring of concrete infrastructure in China and the establishment of a traceable digital operation and maintenance system.

After identifying the YOLOv5 network structure as the original model, the GhostNet backbone network modification, self-attentive mechanism introduction and multi-scale feature fusion are carried out for the original model, resulting in a lightweight crack detection model based on YOLOv5 - YOLOv5-GTB. The improved model not only has a much reduced number of parameters and faster inference speed, but also significantly outperforms the original model in terms of accuracy and mAP, and each improved part has a positive effect on the final results, which basically fulfills the research purpose proposed in this paper.

Reference

[1] 陈玉龙. 高速公路桥梁结构病害与加固措施[J]. 中华建设, 2019(05): 158-159.

[2] 混凝土裂缝深度的激光超声探测方法 - 中国知网[EB/OL]. [2022-05-25]. https://kns.cnki.net/kcms/detail/detail.aspx?dbcode=CJFD&dbname=CJFDLAST2021&filename=ZNGD202103017&uniplatform=NZKPT&v=qe3cK7QlICDd9IC3I2RKYampOKZ71ztbA7yooxVk8y6JChKhGi-CrbnOq2BUZHjj.

[3] LIONG S T, GAN Y S, HUANG Y C, 等. Automatic defect segmentation on leather with deep learning[J]. arXiv preprint arXiv:1903.12139, 2019.

[4] ZHANG L, YANG F, DANIEL ZHANG Y, 等. Road crack detection using deep convolutional neural network[C/OL]//2016 IEEE International Conference on Image Processing (ICIP). 2016: 3708-3712. https://doi.org/10.1109/ICIP.2016.7533052.

[5] ZOU Q, CAO Y, LI Q, 等. CrackTree: Automatic crack detection from pavement images[J]. Pattern Recognition Letters, 2012, 33(3): 227-238.

[6] DUNG C V. Autonomous concrete crack detection using deep fully convolutional neural network[J]. Automation in Construction, 2019, 99: 52-58.

[7] 蔡天池, 刘春, 周骁腾, 等. 基于无人机的建筑立面裂缝检测[J]. 工程勘察, 2022, 50(04): 45-51.

[8] 黄建平. 基于二维图像和深度信息的路面裂缝检测关键技术研究[D/OL]. 哈尔滨工业大学, 2013[2022-05-25]. https://kns.cnki.net/kcms/detail/detail.aspx?dbcode=CDFD&dbname=CDFDLAST2015&filename=1014080650.nh&uniplatform=NZKPT&v=42QFoXiVjIWjT5ZlKLwxieYhjLfNPrGLipia6O9sfAWkhdMOUDhlbxudwrMKoil2.

[9] CHA Y J, CHOI W, SUH G, 等. Autonomous Structural Visual Inspection Using Region-Based Deep Learning for Detecting Multiple Damage Types[J/OL]. Computer-Aided Civil and Infrastructure Engineering, 2018, 33(9): 731-747. https://doi.org/10.1111/mice.12334.

[10] MOHAN A, POOBAL S. Crack detection using image processing: A critical review and analysis[J]. Alexandria Engineering Journal, 2018, 57(2): 787-798.

[11] ZHOU S, BI Y, WEI X, 等. Automated detection and classification of spilled loads on freeways based on improved YOLO network[J/OL]. Machine Vision and Applications, 2021, 32(2): 44. https://doi.org/10.1007/s00138-021-01171-z.

[12] GIRSHICK R, DONAHUE J, DARRELL T, 等. Rich feature hierarchies for accurate object detection and semantic segmentation[C]//Proceedings of the IEEE conference on computer vision and pattern recognition. 2014: 580-587.

[13] LI R, YUAN Y, ZHANG W, 等. Unified vision-based methodology for simultaneous concrete defect detection and geolocalization[J]. Computer-Aided Civil and Infrastructure Engineering, 2018, 33(7): 527-544.

[14] HUYAN J, LI W, TIGHE S, 等. Detection of sealed and unsealed cracks with complex backgrounds using deep convolutional neural network[J]. Automation in Construction, 2019, 107: 102946.

[15] DENG J, LU Y, LEE V C S. Concrete crack detection with handwriting script interferences using faster region-based convolutional neural network[J/OL]. Computer-Aided Civil and Infrastructure Engineering, 2020, 35(4): 373-388. https://doi.org/10.1111/mice.12497.

[16] CUBERO-FERNANDEZ A, RODRIGUEZ-LOZANO F, VILLATORO R, 等. Efficient pavement crack detection and classification[J]. EURASIP Journal on Image and Video Processing, 2017, 2017(1): 1-11.

[17] SHENG P, CHEN L, TIAN J. Learning-based road crack detection using gradient boost decision tree[C]//2018 13th IEEE Conference on Industrial Electronics and Applications (ICIEA). IEEE, 2018: 1228-1232.

[18] HE K, ZHANG X, REN S, 等. Deep residual learning for image recognition[C]//Proceedings of the IEEE conference on computer vision and pattern recognition. 2016: 770-778.

[19] HE K, GKIOXARI G, DOLLÁR P, 等. Mask r-cnn[C]//Proceedings of the IEEE international conference on computer vision. 2017: 2961-2969.

[20] CHEN L C, PAPANDREOU G, KOKKINOS I, 等. Semantic image segmentation with deep convolutional nets and fully connected crfs[J]. arXiv preprint arXiv:1412.7062, 2014.

[21] SUNG K K, POGGIO T. Example-based learning for view-based human face detection[J]. IEEE Transactions on pattern analysis and machine intelligence, 1998, 20(1): 39-51.

[22] DOLLAR P, WOJEK C, SCHIELE B, 等. Pedestrian detection: An evaluation of the state of the art[J]. IEEE transactions on pattern analysis and machine intelligence, 2011, 34(4): 743-761.

[23] KOBATAKE H, YOSHINAGA Y. Detection of spicules on mammogram based on skeleton analysis[J]. IEEE Transactions on Medical Imaging, 1996, 15(3): 235-245.

[24] YU Y, ZHANG J, HUANG Y, 等. Object detection by context and boosted HOG-LBP[C]//ECCV workshop on PASCAL VOC. 2010.

[25] KRIZHEVSKY A, SUTSKEVER I, HINTON G E. ImageNet Classification with Deep Convolutional Neural Networks[C/OL]//Advances in Neural Information Processing Systems: 卷 25. Curran Associates, Inc., 2012[2022-02-13]. https://proceedings.neurips.cc/paper/2012/hash/c399862d3b9d6b76c8436e924a68c45b-Abstract.html.

[26] VEDALDI A, GULSHAN V, VARMA M, 等. Multiple kernels for object detection[C]//2009 IEEE 12th international conference on computer vision. IEEE, 2009: 606-613.

[27] VIOLA P, JONES M. Rapid object detection using a boosted cascade of simple features[C]//Proceedings of the 2001 IEEE computer society conference on computer vision and pattern recognition. CVPR 2001: 卷 1. Ieee, 2001: I-I.

[28] HARZALLAH H, JURIE F, SCHMID C. Combining efficient object localization and image classification[C]//2009 IEEE 12th international conference on computer vision. IEEE, 2009: 237-244.

[29] DALAL N, TRIGGS B. Histograms of oriented gradients for human detection[C]//2005 IEEE computer society conference on computer vision and pattern recognition (CVPR'05): 卷 1. Ieee, 2005: 886-893.

[30] VIOLA P, JONES M J. Robust real-time face detection[J]. International journal of computer vision, 2004, 57(2): 137-154.

[31] LOWE D G. Object recognition from local scale-invariant features[C]//Proceedings of the seventh IEEE international conference on computer vision: 卷 2. Ieee, 1999: 1150-1157.

[32] LIENHART R, MAYDT J. An extended set of haar-like features for rapid object detection[C]//Proceedings. international conference on image processing: 卷 1. IEEE, 2002: I-I.

[33] BAY H, TUYTELAARS T, GOOL L V. Surf: Speeded up robust features[C]//European conference on computer vision. Springer, 2006: 404-417.

[34] FREUND Y, SCHAPIRE R E. Experiments with a new boosting algorithm[C]//icml: 卷 96. Citeseer, 1996: 148-156.

[35] EVERINGHAM M, VAN GOOL L, WILLIAMS C K I, 等. The Pascal Visual Object Classes (VOC) Challenge[J/OL]. International Journal of Computer Vision, 2010, 88(2): 303-338. https://doi.org/10.1007/s11263-009-0275-4.

[36] FELZENSZWALB P, GIRSHICK R, MCALLESTER D, 等. Discriminatively trained mixtures of deformable part models[J]. PASCAL VOC Challenge, 2008.

[37] IANDOLA F N, HAN S, MOSKEWICZ M W, 等. SqueezeNet: AlexNet-level accuracy with 50x fewer parameters and <0.5MB model size[J/OL]. arXiv:1602.07360 [cs], 2016[2022-05-07]. http://arxiv.org/abs/1602.07360.

[38] WOO S, PARK J, LEE J Y, 等. CBAM: Convolutional Block Attention Module[M/OL]//FERRARI V, HEBERT M, SMINCHISESCU C, 等. Computer Vision – ECCV 2018: 卷 11211. Cham: Springer International Publishing, 2018: 3-19[2022-05-24]. http://link.springer.com/10.1007/978-3-030-01234-2_1.

[39] HOU Q, ZHOU D, FENG J. Coordinate Attention for Efficient Mobile Network Design[C/OL]//2021 IEEE/CVF Conference on Computer Vision and Pattern Recognition

(CVPR). Nashville, TN, USA: IEEE, 2021: 13708-13717[2022-05-24]. https://ieeexplore.ieee.org/document/9577301/.

[40] DEVLIN J, CHANG M W, LEE K, 等. Bert: Pre-training of deep bidirectional transformers for language understanding[J]. arXiv preprint arXiv:1810.04805, 2018.

[41] ZHANG M, LI J. A commentary of GPT-3 in MIT Technology Review 2021[J]. Fundamental Research, 2021, 1(6): 831-833.

[42] VÄRTINEN S. Generating Role-Playing Game Quest Descriptions With the GPT-2 Language Model[J]. 2022.

[43] DOSOVITSKIY A, BEYER L, KOLESNIKOV A, 等. An image is worth 16x16 words: Transformers for image recognition at scale[J]. arXiv preprint arXiv:2010.11929, 2020.

[44] CORDONNIER J B, LOUKAS A, JAGGI M. On the relationship between self-attention and convolutional layers[J]. arXiv preprint arXiv:1911.03584, 2019.

[45] HAO W, ZHILI S. Improved Mosaic: Algorithms for more Complex Images[J/OL]. Journal of Physics: Conference Series, 2020, 1684(1): 012094. https://doi.org/10.1088/1742-6596/1684/1/012094.

[46] VASWANI A, SHAZEER N, PARMAR N, 等. Attention is All you Need[C/OL]//Advances in Neural Information Processing Systems: 卷 30. Curran Associates, Inc., 2017[2022-05-24]. https://proceedings.neurips.cc/paper/2017/hash/3f5ee243547dee91fbd053c1c4a845aa-Abstract.html.

[47]TAN M, PANG R, LE Q V. EfficientDet: Scalable and Efficient Object Detection: arXiv:1911.09070[R/OL]. arXiv, 2020[2022-05-24]. http://arxiv.org/abs/1911.09070.

# Appendix A YOLOv5s-GTB feature maps

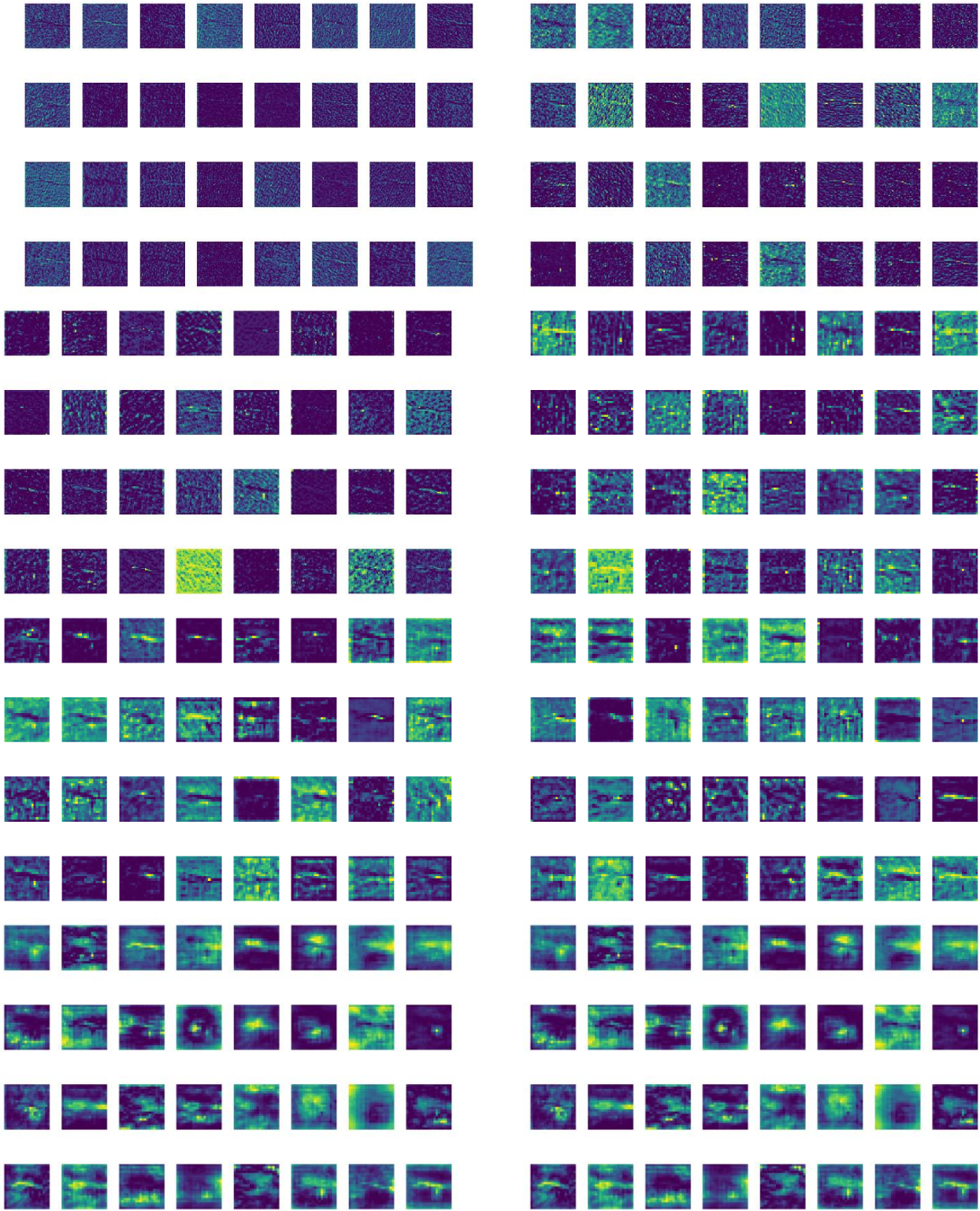

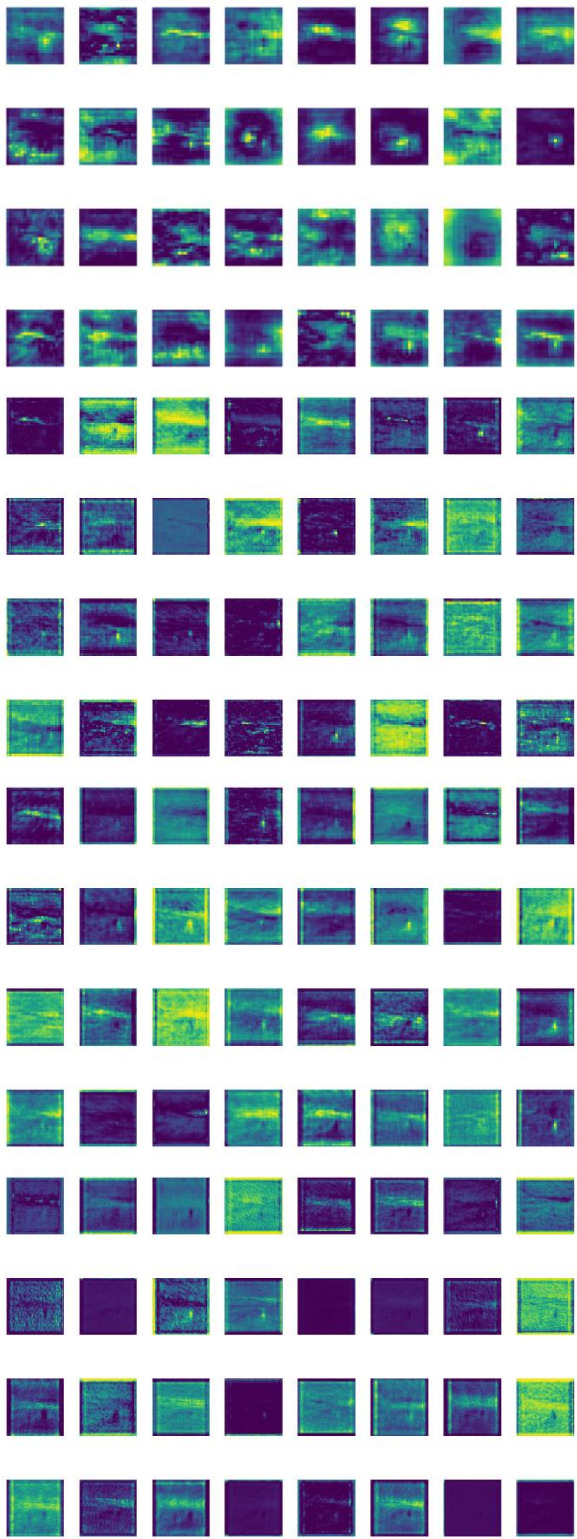
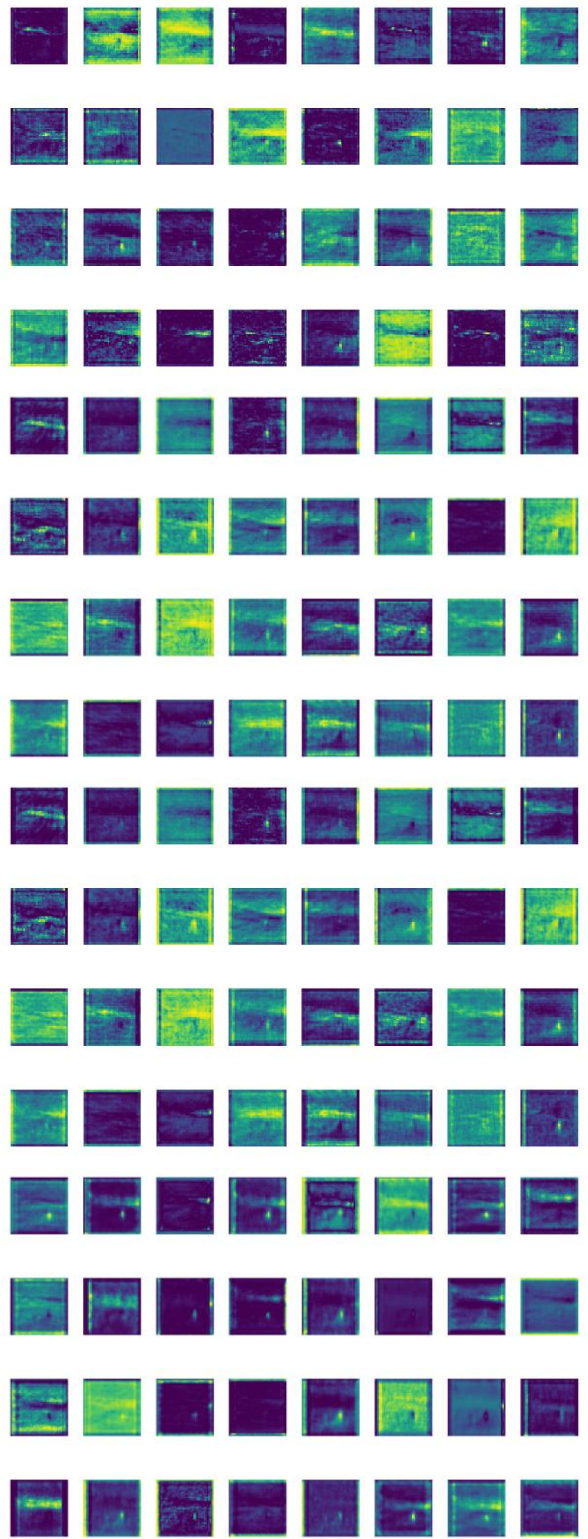

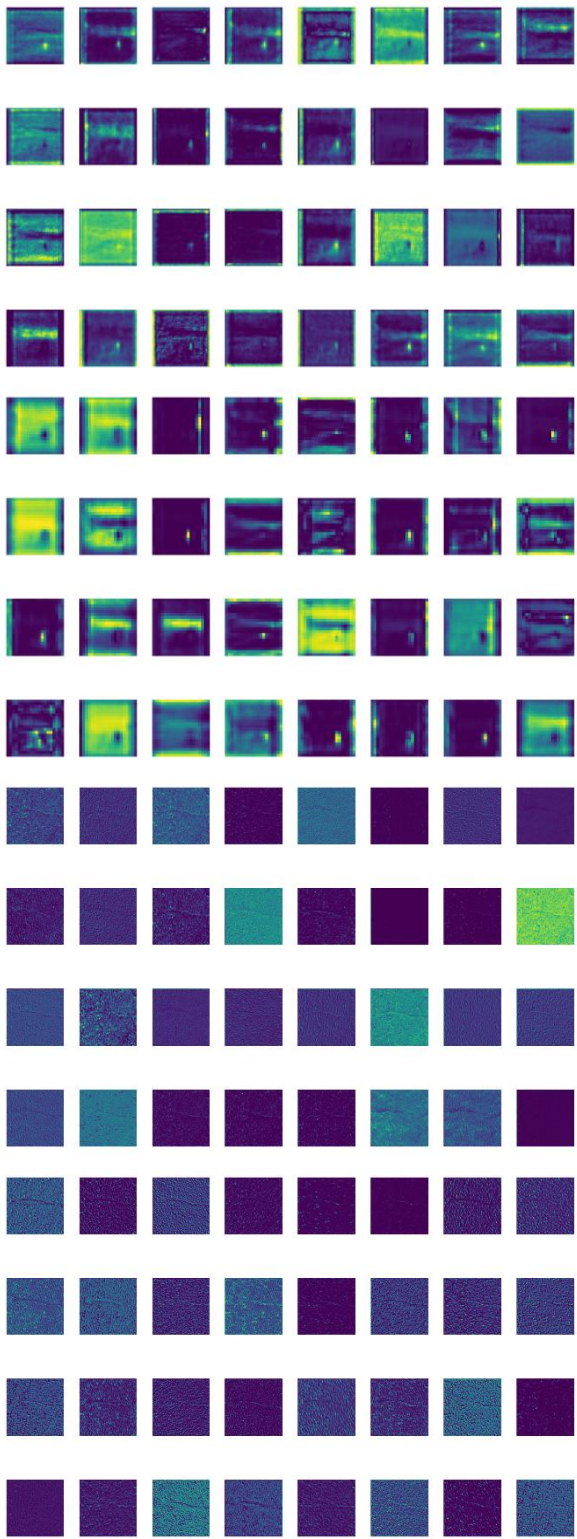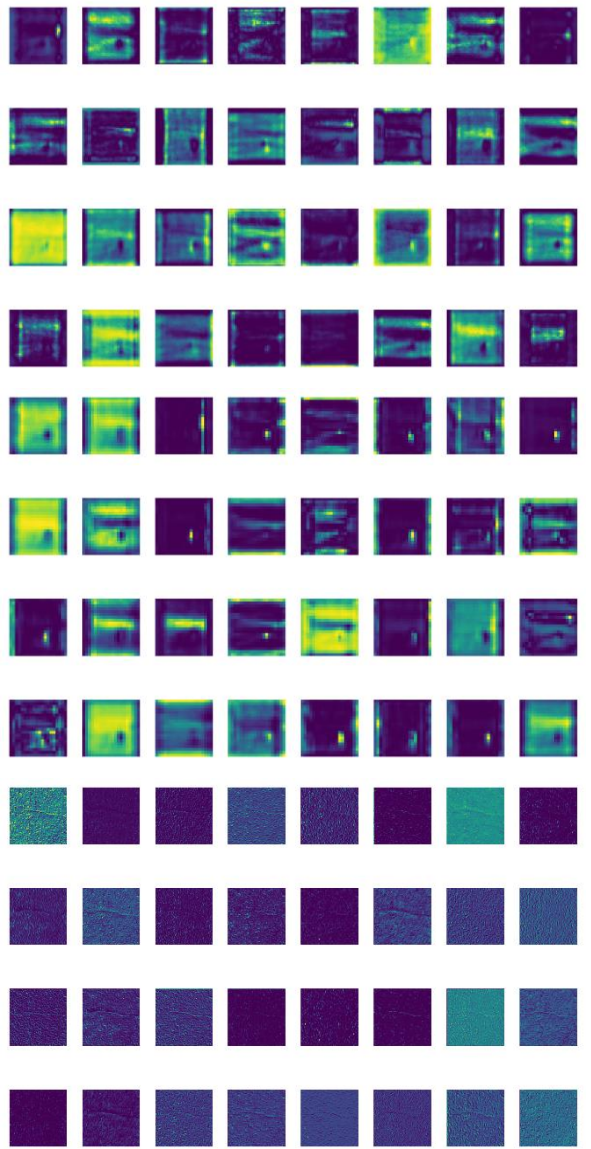